\documentclass[10pt,twocolumn,letterpaper]{article}

\usepackage{cvpr}
\usepackage{times}
\usepackage{epsfig}
\usepackage{graphicx}
\usepackage{amsmath}
\usepackage{amssymb}
\usepackage{color}
\usepackage{multirow}
\usepackage{subfigure}
\usepackage{underscore}

\usepackage[breaklinks=true,bookmarks=false]{hyperref}

\cvprfinalcopy 

\def\cvprPaperID{4718} 

\begin{document}

\title{Bi-Adversarial Auto-Encoder for Zero-Shot Learning}

\author{Yunlong Yu$^1$, Zhong Ji$^1$, Yanwei Pang$^1$, Jichang Guo$^1$, Zhongfei Zhang$^2$, and Fei Wu$^3$\\
$^1$Tianjin University, $^2$Binghamton University, $^3$Zhejiang University
}


\maketitle

\begin{abstract}
   Existing generative Zero-Shot Learning (ZSL) methods only consider the unidirectional alignment from the class semantics to the visual features while ignoring the alignment from the visual features to the class semantics, which fails to construct the visual-semantic interactions well. In this paper, we propose to synthesize visual features based on an auto-encoder framework paired with bi-adversarial networks respectively for visual and semantic modalities to reinforce the visual-semantic interactions with a bidirectional alignment, which ensures the synthesized visual features to fit the real visual distribution and to be highly related to the semantics. The encoder aims at synthesizing real-like visual features while the decoder forces both the real and the synthesized visual features to be more related to the class semantics. To further capture the discriminative information of the synthesized visual features, both the real and synthesized visual features are forced to be classified into the correct classes via a classification network. Experimental results on four benchmark datasets show that the proposed approach is particularly competitive on both the traditional ZSL and the generalized ZSL tasks.
\end{abstract}

\section{Introduction}

In recent years, the deep learning techniques have achieved remarkable performances in both computer vision and machine learning areas, constantly pushing the boundaries of what is possible. The progress partly relies on the growing availability of big data. However, in some cases, the data are difficult to collect, e.g., fine-grained classification data. In order to build powerful models in these problematic situations, Zero-Shot Learning (ZSL) \cite{akata2016label,akata2015evaluation,annadani2018Preserving,changpinyo2016synthesized,lampert2014attribute,yu2018stacked} algorithms have been developed and proven to be a promising direction in the missing data scenarios. The task of ZSL requires classifying the unseen classes that have no visual data available for training, which is achieved by transferring the knowledge from the seen classes to the unseen ones with some semantic information, e.g., attributes \cite{lampert2014attribute} and word vectors \cite{mikolov2013distributed}.

Recently, to address the data missing issue of unseen classes, some approaches \cite{arora2018Generalized,chen2018zero,long2017zero,xian2018feature,zhu2018generative} try to synthesize pseudo visual features for unseen classes with the generative models. In essence, these approaches take as input either the class semantics prototypes or together with some noises to learn a model to narrow down the distribution differences between the synthesized and the real visual features. Once obtained the class semantics prototypes of any unseen classes, the learned model may synthesize the corresponding pseudo visual features as many as possible. However, recent generative zero-shot approaches mostly focus on capturing the visual distribution information via a unidirectional alignment from the class semantics to the visual features, which cannot ensure that the synthesized visual features are semantics-related and discriminative enough to be classified.

\begin{figure*}
\begin{center}
   \includegraphics[height=5.8cm,width=0.75\linewidth]{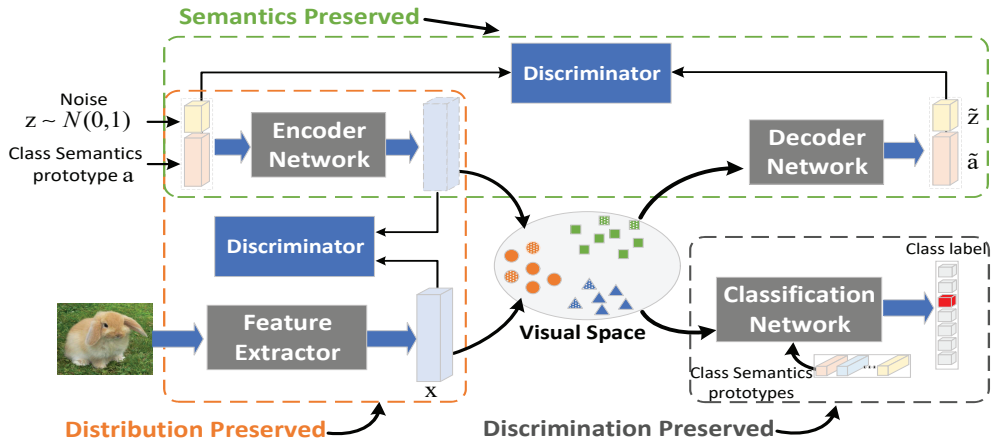}
\end{center}
   \caption{Architecture of the proposed framework. The dashed samples in the visual space represent the generative visual features.}
\label{fig:fig1}
\end{figure*}

To alleviate the above issues, we propose to further explore the visual space from the following two aspects. First, we regularize the synthesized visual features to be highly related to the class semantics by enforcing both the real and synthesized visual features to be well inferred back to the class semantics. Second, we regularize both the real and synthesized visual features to be classified into the ground-truth class labels to capture the discriminative information. Specifically, we propose an auto-encoder framework paired with two respective adversarial networks. The encoder, acting as the visual feature generator, aims at capturing the real visual distribution by formulating the synthesized and the real visual features into an adversarial network. The decoder, acting as the semantics inference that forces both the real and the synthesized pseudo visual features to be related to the class semantics by formulating the inferred and the real class semantics into another adversarial network. We also add a classification network to classify both the real and the synthesized visual features into the correct classes, which encourages the synthesized visual features as much discriminative as the real visual features. The whole framework of the proposed model is illustrated in Fig.~\ref{fig:fig1}.

Consequently, the decoder network and the classification network help the encoder network to boost the feature synthesis by enforcing the synthesized visual features to be semantically related and discriminative, respectively. Compared with the existing generative approaches, this architecture is a bidirectional visual-semantic alignment constraint, which facilitates the interactions between the visual and the class semantics modalities and captures the discriminative information derived from the visual feature space.

It is worthwhile to highlight several aspects of the proposed approach here:
\begin{enumerate}
  \item We propose a generative approach for ZSL based on an encoder-decoder framework to reinforce the visual-semantic interactions with a bidirectional alignment.

  \item We propose two adversarial networks respectively for visual and class semantics modalities to guide the synthesized visual features to fit the real visual feature distribution and to be highly related to the class semantics.

  \item To further capture the discriminative information of the synthesized visual features, we also design a classification network to take as input both the real and synthesized visual features to predict the ground-truth classes, which ensures the synthesized visual features as much discriminative as the real visual features.
\end{enumerate}

We conduct experiments on four benchmark datasets for both traditional ZSL and generalized ZSL tasks. Experimental results show that our proposed approach achieves significant improvements for the traditional ZSL task and achieves better competitive performances for the generalized ZSL \cite{chao2016empirical,Fu2016Semi} task than the state-of-the-art approaches.

\section{Related work}

From the view of constructing the visual-semantic interactions, the existing ZSL approaches could be divided into two categories: the discriminative models and the generative models.
\subsection{Discriminative Models for ZSL}
A simple approach to build visual-semantic interactions is to project the visual features to the class semantics space with a linear \cite{lazaridou2014wampimuk} or a non-linear model \cite{Li2018Discriminative,Morgado2017Semantically,socher2013zero}. Some approaches propose to learn a compatible matrix to obtain the compatibility scores of the visual features and the class semantics prototypes with different objective functions, e.g., ranking loss formulation \cite{akata2016label,frome2013devise}, structural SVM loss \cite{akata2015evaluation}, and the square loss function \cite{romera2015embarrassingly, yu2018transductive}. Recently, as one of the most related efforts to ours, SAE \cite{kodirov2017semantic} employs a linear semantic encoder-decoder framework to regularize the model by enforcing the encoder parameters and the decoder parameters to be symmetric. Although the visual samples are represented with deep features, they cannot effectively handle the semantic inconsistency between the visual and the class semantics modalities, and commonly suffer from the information degradation issue caused by ``heterogeneity gap''. In this work, we propose an encoder-decoder framework paired with the adversarial networks to jointly capture the distribution information of the synthesized visual features and reinforce the visual-semantic alignment.

\subsection{Generative Models for ZSL}
To capture more distribution information from visual space, recent work focuses on generating pseudo features for unseen classes with class semantics prototypes. A simple approach to generating visual features is directly to take as input the class semantics prototype with a linear \cite{dinu2014improving} or a deep model \cite{zhang2017learning}. Compared with the models that project the visual features to the class semantics space, the reversed projection models have the potential to alleviate the hubness issue where some unseen class prototypes (``hub'') tend to appear in the top neighbors of many test instances. We refer readers to \cite{dinu2014improving} for more details. Although promising results have been achieved, these approaches are hard to align the visual and class semantics spaces well since each class has many visual samples in the visual space but only has one class semantics prototype in the class semantics space.

In recent years, significant progress in the generative approaches suggests yielding the desired distribution with a simple instance via functional approximators. Motivated by this idea, some models are proposed to generate pseudo samples for unseen classes with adversarial networks \cite{chen2018zero,tong2018adversarial,xian2018feature,zhu2018generative} and variational auto-encoder \cite{wang2018zero}. Our work is close to \cite{bucher2017generating} in which an adversarial auto-encoder \cite{Makhzani2015Adversarial} is applied for generating visual features. Different with  \cite{bucher2017generating} that employs an adversarial criterion to constrain the latent codes produced by visual features to fit a prior noise distribution, our model reinforces the visual-semantics alignment by employing two adversarial networks to respectively fit the visual distribution and the class semantics distribution.

\section{Approach}
In this section, we first introduce the problem formulation and then discuss in detail the proposed generative model based on the encoder-decoder framework paired with two respective adversarial networks.
\subsection{Problem Formulation}
Given a list of seen samples defined by $N$ triplets $\{\mathbf{x}_i,\mathbf{a}_i,\mathbf{y}_i\}_i^N$, where $\mathbf{x}_i\in\mathbb{R}^p$ is the image feature representation, $\mathbf{a}_i\in\mathbb{R}^q$ is the corresponding class semantics prototype and $\mathbf{y}_i\in\mathcal{Y}_s$ is the associated one-hot class label; $\mathcal{Y}_s$ is the label space of seen classes; $p$ and $q$ are the dimensionalities of the visual and the class semantics spaces, respectively. During the test stage, the unseen class semantics prototypes and the class labels $\{\mathbf{a}_t, \mathbf{y}_t\}$ are provided, where $\mathbf{y}_t\in\mathcal{Y}_t$ and $\mathcal{Y}_s\bigcap\mathcal{Y}_t={\O}$. In the traditional ZSL task, the test sample $\mathbf{x}_t\in\mathbb{R}^p$ comes from unseen classes and is classified into the pre-defined candidate unseen classes $\mathcal{Y}_t$. In contrast, in the generalized ZSL task, the test sample $\mathbf{x}_t$ is either from seen classes or unseen classes and is classified into the set composed of both seen and unseen classes.
\subsection{Bi-Adversarial Auto-Encoder (BAAE)}

In this work, we attempt to synthesize some semantics-related and discriminative samples for unseen classes to address the sample-missing issue in the ZSL task. To this end, we design a generative approach called Bi-Adversarial Auto-Encoder (BAAE) to synthesize visual features for ZSL, as illustrated in  Fig.~\ref{fig:fig1}. In BAAE, two adversarial branches are formulated into an encoder-decoder framework, which separately captures the semantics-related and the visual distribution information. In the encoder branch, the class semantics prototype $\mathbf{a}$ together with the noise vector $\mathbf{z}$ is taken as input to synthesize the pseudo visual feature $\mathbf{\tilde{x}}$ with a generative network, which learns a mapping from a joint space of both the class semantics and noises into the visual space. In the decoder branch, the visual sample $\mathbf{x}$ is decomposed into two independent vectors with an inference network, which learns an inverse mapping from the visual space to the joint space that is spanned by the class semantics and the noise vector.

The adversarial generative model has been employed in some previous approaches \cite{xian2018feature, zhu2018generative}. Different from these methods that mostly focus on synthesizing samples to capture the visual distribution via a unidirectional semantic-visual alignment, we propose to synthesize with a bidirectional alignment, i.e., semantic-visual and visual-semantic alignments, to ensure the synthesized visual features to capture both the semantics-related and feature distribution information. First, the synthesized visual features are forced to be closed to the real visual features to fit the real visual feature distribution. Second, both the real and the synthesized visual features are taken as input to the inference network to infer the corresponding class semantics, which ensures the synthesized visual features to be highly related to the class semantics. Finally, both the real and the synthesized visual features are restricted to be classified into the ground-truth classes, which ensures the synthesized visual features to be discriminative. Inspired by these three points, the objective of the proposed generative approach can be formulated as:
\begin{equation}
  Obj = \mathcal{F}_{encoder} + \mathcal{F}_{decoder} + \mathcal{F}_{cls}.
  \label{equ:equ1}
\end{equation}

For the encoder part, both the class semantics and the noises are concatenated into a holistic vector for the generator network to synthesize pseudo visual features, which is supervised with the real image visual features:
\begin{equation}\label{equ:equ2}
  \mathcal{F}_{align} = \min_\theta \sum_i\|\mathbf{x}_i -\mathbf{\tilde{x}}_i\|_2^2,
\end{equation}
where $\mathbf{\tilde{x}}_i=G_\theta(\mathbf{a}_i,\mathbf{z})$ is the pseudo visual feature synthesized with the corresponding class semantics prototype $\mathbf{a}_i$ and a random Gaussian noise vector $\mathbf{z}$; $\theta$ is the parameter of the generative network $G$. This term encourages that the synthesized visual features are similar to the real visual features.
\begin{figure}
\begin{center}
   \includegraphics[width=0.96\linewidth]{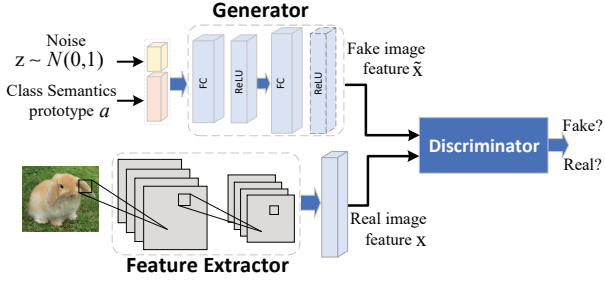}
\end{center}
   \caption{The adversarial process of the visual modality.}
\label{fig:fig2}
\end{figure}
As the visual features are high-level representation, typical reconstruction metrics such as $\ell_p$-norm is hard to capture the visual distribution. To this end, we further proceed both the real visual features and the synthesized pseudo visual features into an adversarial learning process illustrated in Fig.~\ref{fig:fig2}, in which the generator tries to approximate the real-like data distribution while the discriminator is to distinguish whether the features are drawn from the generator's output or the real data distribution:
\begin{equation}\label{equ:equ3}
\begin{aligned}
\mathcal{F}_{adv}=~&\mathbb{E}_{\mathbf{x}\sim{p(\mathbf{x})}}[\log D_\phi(\mathbf{x})] + \\
   &\mathbb{E}_{\mathbf{\tilde{x}}\sim{p_\theta}(\mathbf{\tilde{x}}|\mathbf{z},\mathbf{a})}[\log(1- D_\phi(\mathbf{\tilde{x}}))]
  + \gamma L_{GP},
\end{aligned}
\end{equation}
where $\phi$ is the parameter of the discriminator $D$; $L_{GP}=(\|\nabla_{\mathbf{\hat{x}}}D_\phi(\mathbf{\hat{x}})\|_2^2-1)^2$ is the gradient penalty to enforce the Lipschitz constraint; $\mathbf{\hat{x}}$ is the linear interpolation between the real feature $\mathbf{x}$ and the synthesized feature $\mathbf{\tilde{x}}$; $\gamma$ is a hyperparameter.

For the decoder part, the input is either the real or the synthesized visual features to the semantics inference network that decomposes the input into two separate vectors with two respective subnetworks. One is supervised by the real class semantics and the other is constrained into the noise space. Specifically, the inference network is written as $E_\upsilon(\mathbf{x})\rightarrow[~\mathbf{\tilde{a}};~\mathbf{\tilde{z}}~]$, where $\upsilon$ is the parameter of the inference network $E$. Intuitively, the inferred class semantic vector $\mathbf{\tilde{a}}$ should be close to the real class semantic prototype, i.e.,
\begin{equation}\label{equ:equ4}
  \mathcal{F}_{align}^{'} = \min_{\theta,\upsilon} \sum_i\|\mathbf{a}_i -\mathbf{\tilde{a}}_i\|_2^2.
\end{equation}
Since the samples from the same class share the same class semantic prototype, minimizing Eq.~(\ref{equ:equ4}) encourages to the synthesize visual features from the same class to gather together and capture the class semantics.

\begin{figure}
\begin{center}
   \includegraphics[width=0.95\linewidth]{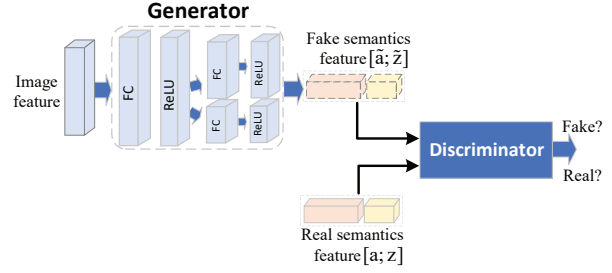}
\end{center}
   \caption{The adversarial process of the class semantics modality. The image feature is either the real visual feature or the synthesized pseudo visual feature.}
\label{fig:fig3}
\end{figure}

Just as the visual features, the class semantics are high-level representations; the euclidean distance is hard to capture semantics information. Hence, we also adopt adversarial learning for the semantics inference, as illustrated in Fig.~\ref{fig:fig3}. Specifically, the decoder network is seen as the generative network of the adversarial process. A discriminator is designed to distinguish whether the input is from the generator's output or the real data distribution. The real data are the concatenation of the class semantic vector and the random Gaussian noise vector, i.e., the input of the encoder network. Similar to Eq.~(\ref{equ:equ3}), the adversarial process is formulated as:
\begin{equation}\label{equ:equ5}
\begin{aligned}
\mathcal{F}_{adv}^{'}=~&\mathbb{E}_{(\mathbf{a},\mathbf{z})\sim{p(\mathbf{a},\mathbf{z})}}[\log D_\omega([\mathbf{a};\mathbf{z}])] + \\
   &\mathbb{E}_{(\mathbf{\tilde{a}},\mathbf{\tilde{z}})\sim{p_\upsilon}
   (\mathbf{\tilde{a}},\mathbf{\tilde{z}}|\mathbf{x}),
   p_\upsilon(\mathbf{\tilde{a}},\mathbf{\tilde{z}}|\mathbf{\tilde{x}})p_\theta(\mathbf{\tilde{x}}|[\mathbf{a},\mathbf{z}])}
   [\log(1- D_\omega([\mathbf{\tilde{a}};\mathbf{\tilde{z}}]))]\\
  &+ \eta L_{GP},
\end{aligned}
\end{equation}
where $\omega$ is the parameter of the discriminator; $L_{GP}$ is the gradient penalty, $[\cdot;\cdot]$ is the concatenation operator, $\eta$ is the hyperparameter.
As mentioned above, the whole network is achieved as a closed loop, in which the visual-semantic interaction is reinforced with a bidirectional alignment. However, a robust visual-semantic interaction cannot derive the discriminative power of the synthesized visual features, which is vital for the classification. To boost the discriminative power of the synthesized visual features, we design a classification network to take as input the real and the synthesized visual features to predict the corresponding class labels, which is formulated as:
\begin{equation}\label{equ:equ6}
  \mathcal{F}_{cls} = \min_{\psi,\theta}\sum_i(L_\psi(\mathbf{x}_i,\mathbf{A})+L_\psi(\mathbf{\tilde{x}}_i,\mathbf{A})),
\end{equation}
where $L_\psi(\mathbf{x}_i,\mathbf{A})$ and $L_\psi(\mathbf{\tilde{x}}_i,\mathbf{A})$ are the classification losses of real and synthesized pseudo visual features, respectively. $\psi$ is the parameter of the classification network. This term encourages the synthesized visual features as much discriminative as the real visual features to be classified into the ground-truth classes. Specifically,
\begin{equation}\label{equ:equ7}
  L_\psi(\mathbf{x}_i,\mathbf{A}) = -\log P(\mathbf{y}|\mathbf{x}_i,\mathbf{A};\psi),
\end{equation}
where $\mathbf{A}\in\mathbb{R}^{q\times{M}}$ is the class semantics prototype matrix of both the seen and unseen classes, and $M$ is the number of all classes. $P(y_j|\mathbf{x}_i,\mathbf{A};\psi)=\frac{exp(\mathbf{x}_i^TF_\psi(\mathbf{a}_j))}{\sum_k^Mexp(\mathbf{x}^TF_\psi(\mathbf{a}_k))}$, where $\mathbf{a}_k\in\mathbf{A}$; $\mathbf{a}_j$ is the corresponding class semantics prototype of class $\mathbf{y}_j$; $F_\psi$ is the linear function to project the class semantics into the visual space. The value of $\mathbf{x}_i^TF_\psi(\mathbf{a}_j)$ is seen as the compatibility score between the visual feature $\mathbf{x}_i$ and the $j$-th class semantic prototype $\mathbf{a}_j$. If the sample $\mathbf{x}_i$ belongs to class $\mathbf{y}_j$, their compatibility score should be large; otherwise it should be small. In this way, the separability between any two different classes is enlarged. Besides, the unseen class semantic prototypes are also taken into consideration, which prevents the seen data from classifying into unseen classes. The seen to unseen bias issue thus is mitigated obviously.

Overall, the objective function of the proposed model is summarized with:
\begin{equation}
  Obj = \mathcal{F}_{align}+\mathcal{F}_{align}^{'}+\mathcal{F}_{adv}+\mathcal{F}_{adv}^{'}+\lambda\mathcal{F}_{cls}
  +\mu R(\theta,\upsilon),
  \label{equ:equ8}
\end{equation}
where $R(\theta,\upsilon)$ is the regulairizer on the parameters; $\lambda$ and $\mu$ are two balance scalars.

\subsection{Apply BAAE for ZSL}
With the proposed BAAE, each unseen class can synthesize its corresponding pseudo visual features in the visual space with the provided class semantic prototype. During the test stage, the similarities between the test instance and the unseen class semantics prototypes are obtained by calculating the distances of the visual features and the synthesized unseen pseudo visual features. In this way, the test instance is classified with the Nearest Neighbor (NN) classifier based on the distances. Furthermore, each class may obtain a lot of pseudo visual features with different noise inputs, and the classification is also achieved by training a parametric classifier, e.g., softmax or SVM.

\section{Experiments}
\begin{table}\label{lab:lab1}
\begin{center}
\begin{tabular}{|l|c|c|c|c|}
\hline
Dataset &Attribute &Image & $\sharp{\mathcal{Y}}_s$ & $\sharp{\mathcal{Y}}_t$\\
\hline\hline
AwA1 &85 & 30,475 & 40 & 10\\
AwA2 &85 & 37,322 & 40 & 10\\
aPY  &64 &15,339 &20  & 12\\
SUN  &102 &14,340 &645 & 72\\
\hline
\end{tabular}
\end{center}
\caption{The statistics of the four benchmark datasets. $\sharp{\mathcal{Y}}_s$ and $\sharp{\mathcal{Y}}_t$ represent the number of seen and unseen classes, respectively.}
\label{table:table1}
\end{table}

In this section, we first document the datasets and experimental settings. Then we present the comparison results of the proposed model on both traditional ZSL and generalized ZSL tasks. Finally, we discuss the impacts of both the classifiers and the number of synthesized visual samples on the proposed generative model.\\
\textbf{Datasets.} We conduct experiments on four benchmark datasets: AwA1 \cite{lampert2014attribute}, AwA2 \cite{xian2017zero}, aPY \cite{farhadi2009describing}, and SUN \cite{patterson2012sun}. These datasets are all annotated with attributes that are used as the class semantics prototypes. The statistics of the datasets are listed in Table \ref{table:table1}.\\
\textbf{Features.} As the visual representations, we use the features released by \cite{xian2017zero}, which are extracted as 2048-dim top layer pooling units of the 101-layered ResNet. The visual features are scaled to [0,~1] with normalization. For the class semantics prototypes, we use the attributes provided by the datasets. Specifically, for both AwA1 and AwA2 datasets, we use the class-level attributes directly and average the image-level attributes to represent the class semantics prototypes for both aPY and SUN datasets. \\
\subsection{Results of the traditional ZSL}
\begin{table}
\begin{center}
\begin{tabular}{|l|c|c|c|c|}
\hline
Method & AwA1 & AwA2 & aPY & SUN\\
\hline\hline
DAP \cite{lampert2014attribute} &44.1 & 46.1 & 33.8 & 39.9\\
SSE  \cite{zhang2015zero} &60.1 & 61.0 & 34.0 & 51.5\\
LATEM \cite{xian2016latent} &55.1 & 55.8 & 35.2 & 55.3\\
ALE \cite{akata2016label} &59.9 & 62.5 & 39.7 & 58.1\\
DEVISE \cite{frome2013devise} &54.2 & 59.7 & 39.8 & 56.5\\
SJE \cite{akata2015evaluation}&65.6 & 61.9 & 32.9 & 53.7\\
ESZSL \cite{romera2015embarrassingly} &58.2 & 58.6 & 38.3 & 54.5\\
SAE \cite{kodirov2017semantic}  &53.0 & 54.1 & 8.3 & 40.3\\
GFZSL \cite{verma2017simple} &  68.3 & 63.8 & 38.4 & 60.6\\
DEM \cite{zhang2017learning}& 68.4 & 67.1 & 40.9 & 61.4\\
RELATION NET \cite{yang2018learning} & 68.2 & 64.2 & - &-\\
GAZSL \cite{zhu2018generative} &68.2 &69.0 &41.1 &61.3\\
CLSWGAN+SM \cite{xian2018feature} & 68.2 & - & - & \textbf{62.1}\\
\hline\hline
BAAE (Ours) &\textbf{71.4} &\textbf{69.4} &\textbf{43.2}& \textbf{62.1}\\
\hline
\end{tabular}
\end{center}
\caption{The per-class average Top-1 accuracy \textbf{T} (\%) of the traditional ZSL. The best results are marked with boldface. All the comparative approaches are achieved with NN except CLSWGAN that is achieved with softmax classifier (``SM'' for short).}
\label{table:table2}
\end{table}
\textbf{Evaluation protocol.}
For the traditional ZSL task that assumes that the test data all come from unseen classes, we use the average per-class top-1 accuracy \textbf{T} following the most majority of the prior work. For the generalized ZSL task, we compute the average per-class top-1 accuracy \textbf{s} on the seen classes, the average per-class top-1 accuracy \textbf{u} on the unseen classes, and their harmonic mean, i.e. $\textbf{H }= 2 \times(\textbf{s}\times \textbf{u})/(\textbf{s} + \textbf{u})$.\\
\begin{table*}
   \begin{center}
    \begin{tabular}{|l|c|c|c|c|c|c|c|c|c|c|c|c|c|}
    \hline
    \multirow{2}{*}{Method} &\multicolumn{3}{c|}{AwA1} &\multicolumn{3}{c|}{AwA2} &\multicolumn{3}{c|}{aPY} &\multicolumn{3}{c|}{SUN}\\
    \cline{2-13}
    & \textbf{u} & \textbf{s} &\textbf{H} &\textbf{u} & \textbf{s} &\textbf{H} &\textbf{u} & \textbf{s} &\textbf{H } &\textbf{u} & \textbf{s} &\textbf{H}\\
    \hline
    SSE \cite{zhang2015zero}   &7.0 &80.5 &12.9   &8.1 &82.5 &14.8   &0.2 &78.9 &0.4    &2.1 &36.4 &4.0\\
    LATEM \cite{xian2016latent} &7.3 &71.7 &13.3   &11.5 &77.3 &20.0   &0.1 &73.0 &0.2    &14.7 &28.8 &19.5\\
    ALE \cite{akata2016label}  &16.8 &76.1 &27.5  &14.0 &81.8 &23.9   &4.6 &73.7 &8.7    &21.8 &33.1 &26.3\\
    DEVISE \cite{frome2013devise} &13.4 &68.7 &22.4  &17.1 &74.7 &27.8   &4.9 &76.9 &9.2    &16.9 &27.4 &20.9\\
    SJE \cite{akata2015evaluation}  &11.3 &74.6 &19.6  &8.0 &73.9 &14.4    &3.7 &55.7 &6.9    &14.7 &30.5 &19.8\\
    ESZSL \cite{romera2015embarrassingly} &2.4 &70.1 &4.6    &5.9 &77.8 &11.0  &11.0 &27.9 &15.8  &11.0 &27.9 &15.8\\
    SAE \cite{kodirov2017semantic} &1.8 &77.1 &3.5    &1.1 &82.2 &2.2  &0.4 &80.9 &0.9    &8.8 &18.0 &11.8 \\
    GFZSL \cite{verma2017simple} &1.8 &80.3 &3.5    &2.5 &80.1 &4.8    &0.0 &\textbf{83.3} &0.0    &0.0 &39.6 &0.0\\
    RELATION NET \cite{yang2018learning} &31.4 &\textbf{91.3} &46.7 &30.0 &\textbf{93.4} &45.3 &- &- &-  &- &- &-\\
    GAZSL \cite{zhu2018generative} &25.7 &82.0 &39.2 &27.0 &82.4 &40.6 &14.2 &78.6 &24.0 &21.7	&34.5	&26.7\\
    CLSWGAN+SM \cite{xian2018feature} &\textbf{57.9} &61.4 &59.6 &- &- &- &- &- &-  &\textbf{42.6} &36.6 &\textbf{39.4}\\
    \hline
    \hline
    BAAE (Ours)&51.0 &84.3 &\textbf{63.4}  &\textbf{51.4} &85.6 &\textbf{64.2}  &\textbf{15.4} &74.1 &\textbf{25.5}   &23.1 &\textbf{36.7} &28.4\\
    \hline
    \end{tabular}
    \end{center}
    \caption{\upshape The performances (in \%) of the generalized ZSL on four datasets. The best results are marked with boldface. All the competitors are achieved with NN except CLSWGAN that is achieved with softmax classifier.}
    \label{table:table3}
\end{table*}
\textbf{Implementation details.}
The proposed BAAE has many parameters, including the hidden layer number, the neuron number of each hidden layer, the hyperparameters, the number of batch size, and the learning rate. In BAAE, both the encoder and the decoder networks have two layers; each layer is activated with the ReLU function. In practice, we have found that the neuron number of the hidden layer is robust to the final performance when it surpasses 500. Thus we set the neuron number of the hidden layer as 1,024 for both the encoder and decoder networks. The remaining parameters are fine-tuned with a cross-validation procedure in which 20\% seen classes are considered as the validation set, allowing to choose the hyperparameters
maximizing the accuracy on the validation set. Specifically, we have found that the proposed BAAE works well when the neuron number of the hidden layer of the discriminator is set as 64. The hyperparameters $\lambda$ and $\mu$ are set 0.01 and 0.001, respectively.

The trained model parameters are initialized with a Gaussian distribution ($\sigma=0.01$) and optimized with the Adam solver with a cross-validated learning rate $0.0001$, using mini-batches of size 48. The model is implemented with the Tensorflow framework running on a Tesla K40 GPU. Given a set of hyperparameters, the training process takes around 10 minutes for each model on AwA1 dataset. Our codes will be released publicly.

First we conduct experiments on the traditional ZSL task and select thirteen approaches for comparison. Specifically, DAP \cite{lampert2014attribute} is one of the pioneering efforts that achieves ZSL with a probabilistic framework to transfer knowledge from the visual features to the class labels using the attributes as intermediary. SSE \cite{zhang2015zero}, LATEM \cite{xian2016latent}, ALE \cite{akata2016label}, DEVISE \cite{frome2013devise}, SJE \cite{akata2015evaluation}, and ESZSL \cite{romera2015embarrassingly} are compatibility based approaches that maximize the compatibility scores of both the visual features and the class semantics prototypes with different objective functions. 
SAE \cite{kodirov2017semantic} is a linear encoder-decoder framework in which the class semantics prototypes serve as the latent representations of the auto-encoder. GFZSL \cite{verma2017simple} and DEM \cite{zhang2017learning} are generative approaches with nonlinear models. To be more specific, GFZSL \cite{verma2017simple} models each class-conditional distribution as an exponential family distribution, while DEM \cite{zhang2017learning} reconstructs the visual samples with two-layer neural networks conditioned on the class semantics prototypes. The results of all the above approaches except DEM \cite{zhang2017learning} are directly cited from \cite{xian2017zero}. RELATION NET \cite{yang2018learning} introduces meta-learning mechanism to learn a deep similarity metric to align both the visual and the class semantics modalities. GAZSL \cite{zhu2018generative} and CLSWGAN \cite{xian2018feature} both equip the Wasserstein GAN \cite{arjovsky2017wasserstein} approaches with discriminative ability to generate pseudo samples for unseen classes. All the competitors use the same features and the same experimental settings as ours. The comparison results are summarized in Table \ref{table:table2}.

From the results in Table \ref{table:table2}, we observe that our BAAE achieves the best performance on four datasets. Specifically, BAAE obtains 3.0\% and 2.1\% improvements over the second best competitors, demonstrating that the proposed architecture learns a more robust visual-semantic alignment for ZSL. Specifically, compared with SAE \cite{kodirov2017semantic}, a similar encoder-decoder framework as our BAAE but with linear model, BAAE obtains significant improvements on four datasets, which indicates the effectiveness of the nonlinear and adversarial models. Compared with CLSWGAN+SM \cite{xian2018feature} that also applies an adversarial network to align the generated distribution and the real visual feature distribution, BAAE also secures better performances, which indicates that both the semantic inference process and the designed regularizers bring positive impacts to the accuracy improvement. Besides, from the results, we observe that the generative approaches i.e., GAZSL \cite{zhu2018generative}, CLSWGAN+SM \cite{xian2018feature}, and the proposed BAAE, perform much better than the other competitors, which indicates the effectiveness of the generative strategies. The reason locates that the synthesized visual features of the generative approaches are more tightly centered around the corresponding real visual distribution, which has the potential to alleviate the ``hubness'' issue.
\begin{table*}
\begin{center}
\begin{tabular}{|c|c|c|c|c|c|c|c|c|c|c|c|c|c|c|c|c|}
    \hline
    \multirow{2}{*}{C} &\multicolumn{4}{c|}{AwA1} &\multicolumn{4}{c|}{AwA2} &\multicolumn{4}{c|}{aPY} &\multicolumn{4}{c|}{SUN}\\
    \cline{2-17}
    &\textbf{T} &\textbf{u} & \textbf{s} &\textbf{H}  &\textbf{T} &\textbf{u} & \textbf{s} &\textbf{H}&\textbf{T} &\textbf{u} & \textbf{s} &\textbf{H} &\textbf{T} &\textbf{u} & \textbf{s} &\textbf{H}\\
    \hline
    NN$\ddag$ &71.4 &\textbf{51.0} &84.3 &\textbf{63.4} &68.6 &\textbf{51.4} &85.6 &\textbf{64.2} &43.2 &\textbf{14.1} &74.1 &\textbf{23.7 } &62.1 &23.1 &36.7 &28.4\\
    NN$^\S$    &71.4 &34.8 &86.2 &49.6 &68.6 &36.1 &86.4&50.9 &43.2 &10.1 &76.4 &17.8 &62.1 &\textbf{21.6} &45.2 &\textbf{29.2}\\
    SM$\ddag$ &\textbf{71.9 }&40.8 &86.1 &55.4 &\textbf{69.4} &42.3 &86.4 &56.8 &\textbf{45.6} &12.6 &78.8 &21.7 &\textbf{63.6} &19.8 &54.3 &29.0\\
    SM$^\S$  &\textbf{71.9} &17.0 &\textbf{89.5}&28.6  &\textbf{69.4} &12.8 &\textbf{89.8 } &22.6 &\textbf{45.6 } &6.6 &\textbf{85.2} &12.3 &\textbf{63.6} &10.2 &\textbf{66.8}&17.7\\
    \hline
\end{tabular}
\end{center}
\caption{\upshape The performances (in \%) of BAAE with different classifiers on four datasets. NN and SM are nearest neighbour classifier and softmax classifier, respectively. $\ddag$ and $\S$ indicate the classifiers with the synthesized pseudo and the ground-truth visual features, respectively.}
\label{table:table4}
\end{table*}

\subsection{Results of the Generalized ZSL}
We then conduct experiments on the generalized ZSL task, and compare our proposed model with eleven  state-of-the-art ZSL approaches in Table \ref{table:table3}. All the results of the competitors are cited from the published papers. From the results in Table \ref{table:table3}, we observe that the proposed BAAE model performs particularly competitive under the more realistic generalized ZSL task. Taking the harmonic mean (\textbf{H}) metric as an example, our BAAE obtains superior results with a large margin against the competitors on AwA1, AwA2 and aPY datasets, and is only outperformed by CLSWGAN+SM \cite{xian2018feature} on SUN dataset. This indicates that the proposed model performs better than the other competitors on alleviating the issue of the seen-unseen bias under the generalized ZSL scenario, which means that the proposed approach can improve the performances of unseen classes while maintaining the seen classes performances. Besides, we observe that the classification performances of the seen classes are much better than those of unseen classes, which indicates that the synthesized pseudo visual features are unlikely to be as good as real visual features.

\subsection{Impacts of classifiers}
As for the seen classes, the classification may be achieved either with the synthesized pseudo visual features or the ground-truth visual features. In this part, we conduct experiments on the four datasets to validate the impacts of different classifiers (i.e., NN and softmax) under different settings on the proposed BAAE. The NN classifier mostly evaluates the discriminative information of the synthesized visual features while the softmax classifier evaluates both the discriminative information and the distribution information of the synthesized visual features. From the experimental results in Table \ref{table:table4}, we have the following observations. (1) The performances \textbf{T} on the traditional ZSL task of softmax classifier are slightly better than those with NN on all datasets, which indicates that the synthesized pseudo visual features are discriminative enough to be classified. (2) The performances \textbf{s} on the seen classes with ground-truth visual features are better than those with the synthesized pseudo visual features, while the performances \textbf{u} on the unseen classes are inferior correspondingly. Besides, the  harmonic mean performances \textbf{H} with generated visual features are more robust than those with ground-truth visual features. The bias indicates that the distribution of the synthesized pseudo visual features is still not as good as the real feature distribution. (3) The harmonic mean performances \textbf{H} with NN classifier are more stable than those with the softmax classifier, which indicates that distribution information of the synthesized visual features is more likely to cause the seen-unseen bias, and the discriminative information contributes more to preserve the performances than the distribution information.

\subsection{Impacts of the synthesized sample number}
\begin{figure}
\begin{center}
   \includegraphics[width=0.95\linewidth]{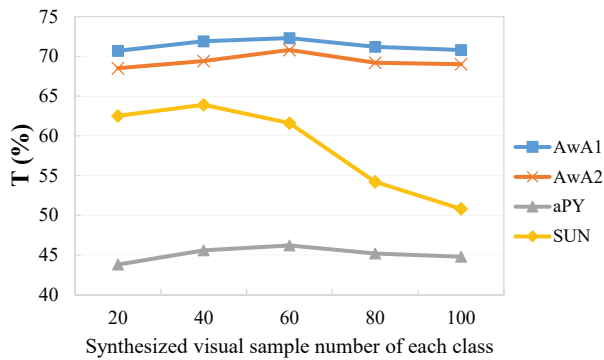}
\end{center}
   \caption{The average per-class top-1 accuracy \textbf{T} of BAAE with different synthesized sample numbers of each unseen class on different datasets.}
\label{fig:fig4}
\end{figure}
In this section, we conduct experiments to evaluate the impacts of the visual distribution for the classification performances. Specifically, we evaluate the average per-class top-1 accuracy \textbf{T} of the proposed BAAE model on the traditional ZSL task via varying the synthesized sample number of each unseen class. As illustrated in Fig.~\ref{fig:fig4}, we observe that the accuracies initially increase and achieve their peaks and then decline with the further increase of the synthesized visual feature number of each unseen class. This indicates that the performances benefit from the visual distribution within a certain range. With the increase of the synthesized visual samples of each class, the distribution information may attenuate the discriminative information. Besides, we observe that the performances on the SUN dataset are much more sensitive than those of the other three datasets. The reason is that the SUN dataset is a fine-grained dataset of which the inter-class differences are small, leading to the fact that discriminative information is easier to be affected by the visual distribution.
\section{Conclusion}

In this paper, we have proposed a novel generative approach for ZSL by synthesizing semantics-related and discriminative visual features. It is based on an auto-encoder framework paired with two respective adversarial networks to fit the real visual distribution and preserve semantics. We also add a classification network to regularize the synthesized visual features to be discriminative. Extensive experimental results show that the proposed approach achieves state-of-the-art performance on the traditional ZSL task and improves a large margin on the generalized ZSL task under the harmonic mean metric.

{\small
\bibliographystyle{ieee}
\bibliography{egbib}
}

\end{document}